\documentclass{spie}

\usepackage{cite}

\usepackage{times} 
\usepackage{indentfirst} 

\usepackage[colorinlistoftodos,color=pink]{todonotes} 
\usepackage{xkeyval}
\presetkeys{todonotes}{inline}{}

\usepackage{mathtools}

\usepackage{enumitem}

\usepackage[hyphens]{url}

\usepackage{multirow}

\DeclarePairedDelimiterX{\norm}[1]{\lVert}{\rVert}{#1}

\title{Linear colour segmentation revisited}

\author{Anna Smagina\supit{1}, Valentina Bozhkova\supit{1}, Sergey Gladilin\supit{1}, Dmitry Nikolaev\supit{1} 		
  \skiplinehalf
  \normalsize 
  \supit{1}Institute for Information Transmission Problems of the Russian Academy of Sciences (Kharkevich Institute)
}

\begin{document}
    \maketitle
    
    \begin{abstract}
    In this work we discuss the known algorithms for linear colour segmentation based on a physical approach and propose a new modification of segmentation algorithm. 
    This algorithm is based on a region adjacency graph framework without a pre-segmentation stage. 
    Proposed edge weight functions are defined from linear image model with normal noise.
    The colour space projective transform is introduced as a novel pre-processing technique for better handling of shadow and highlight areas.
    The resulting algorithm is tested on a benchmark dataset consisting of the images of 19 natural scenes selected from the Barnard's DXC-930 SFU dataset and 12 natural scene images newly published for common use.
    The dataset is provided with pixel-by-pixel ground truth colour segmentation for every image.
    Using this dataset, we show that the proposed algorithm modifications lead to qualitative advantages over other model-based segmentation algorithms, and also show the positive effect of each proposed modification.
    The source code and datasets for this work are available for free access at 
    http://github.com/visillect/segmentation.
\end{abstract}

    \section{Introduction}

Colour segmentation (CS) is one of the most interesting problems in image analysis. 
Its goal is to split images into segments~-- non-intersecting areas corresponding to uniformly coloured objects or their parts. Colour segmentation is used in e.g. augmented reality technology and object tracking in video streams. 
Although the colour of an object in the desired area of the image is considered constant, the colour of pixels in it can vary significantly due to differences in the illumination and the viewing angle for each pixel.
Since segmentation divides an image into areas according to colour of physical world objects and not the colour of pixels, it can be viewed as a special case of the colour constancy problem \cite{gijsenij2011computational}, which is to determine the colour parameters of an object by its image. 
However, the  CS is much easier since it aims only to outline the boundaries of uniformly coloured objects without determining the colour itself. 

The physical approach to the CS \cite{cheng2001color} consists in construction of the algorithms using mathematical models of image formation, which are derived from physical laws of light reflection and scattering. 
Most significantly this approach is presented in Klinker's\cite{klinker1990physical} and Nikolaev's \cite{nikolaev} algorithms. These algorithms are viewed in detail in this paper and we propose an algorithm combining both of them. The image formation model can be divided into an optical image formation model -- the spectral distribution of the sensor (camera) lighting -- and the sensor model that forms a digital image from the optical one.
Each point of the resulting digital image is a vector in the colour space (CSp).  
Usually RGB-CSp is used, so we will consider the CSp three-dimensional. 
The approach based on the linear model of image formation makes it possible to outline a uniformly coloured object with high accuracy by analysing the shape of clusters corresponding to the object in different lighting conditions in the CSp. 

Although the sensor intensity transfer function in most cases is non-linear \cite{klinker1988measurement}, this non-linearity can often be adjusted by the calibration \cite{shepelev2018weighted}. 
If the image source is unknown and no calibration data are available, a blind calibration can be applied similar to the proposed one for a radial distortion  \cite{kunina2016blind}. 
Special CS algorithms which do not assume sensor linearity \cite{kim2004new} are required only when non-linearity correction is impossible, and are not considered in this paper. 

In most recent works about colour \cite{diaz-cortes2017color} as well as semantic and instance segmentation \cite{garcia2017review} problems that we have studied, neural networks are used without image formation model. Classical algorithms may also be used for image segmentation \cite{bertrand2018segmentation, zhang2017unsupervised}, although works studying that are quite rare.
In this work we assume that a more accurate result with less risk of overfitting can be achieved by combining neural-network and classical model-based approaches. 
For example, in work \cite{bai2017deep}  the neural network is used to calculate the potential energy of each pixel in the classical segmentation watershed algorithm.
Another example of combining neural-network and algorithmic approaches can be found in \cite{sheshkus2018vanishing}, where authors propose deep neural networks architecture based on the classical convolutional network, but also containing additional intermediate layers calculating the fast Hough transform. From that perspective studying the physical approach to CS remains significant, despite the active development of machine learning methods.
    \section{HISTORY OF PHYSICS-BASED APPROACH TO LINEAR COLOUR SEGMENTATION}

\subsection{Klinker's algorithm}

One of the first published physics-based CS algorithms was that of Klinker with co-authors \cite{klinker1990physical} based on Shafer’s \cite{shafer1985using} most known model of spectral distribution of sensor illumination -- dichromatic reflectance model (DRM). 
The Schafer’s model describes a large class of materials -- inhomogeneous dielectrics -- which includes paints, plastics, ceramics, paper and different natural materials (see also \cite{lee1990modeling,tominaga1994dichromatic,marszalec2000physics}). 
For a uniformly coloured object covered with glossy inhomogeneous dielectric and one dominating light source the model states that the light reflected by the object can be decomposed into linear combination of specular (interface) and diffuse (body) reflectance components. 
In a sensor CSp such an object generates clusters that have the skewed T- or L-shapes \cite{klinker1990physical,novak1994method}. 
The first stroke of such cluster shape -- body part -- extends from the black point that defines the lower end of this vector to the point of maximum body reflection. 
The other part starts somewhere along the body-reflection cluster and extends to the highlight maximum.

The Klinker's algorithm aims at distinguishing such clusters. 
It consists of two stages: pre-segmentation and main segmentation.
At the pre-segmentation stage the image content is not used, rather, the entire image is divided regularily into square segments (cells) of equal size. 
Then for each segment a principal components analysis (PCA) of its colour distribution is carried out. 
As a result the colour distribution of each segment is classified as pointlike, linear, planar or volumetric. 
From this point onwards, each segment is characterized by its model in the CSp -- a specific point, line or plane, the coordinates of which are set by the center of mass of the colour cluster and its eigenvectors.
In a case of mergers or changes in segment boundaries the model is re-calculated. 

The main segmentation stage starts with the neighbouring segments being checked in pairs for the similarity of their colour model and merged into one if they pass the test. 
According to this test their class should be the same, not to be volumetric and persist after the merge. 
Further processing is based on modelling clusters with a straight line. 
Each of clusters, in descending order of size, attaches neighbouring pixels of other segments that satisfy (according to the distance to cluster's axis) the model of the current segment well enough. 
Segments of other classes can be reduced in area or be fully absorbed.
Since the linear clusters near the zero of the CSp are very close to each other, the processing of pixels close to zero is carried out in a special way. 
Then, if the L-shape test for clusters of neighbouring segments indicates the presence of a highlight, the linear segments merge into planar according to the DRM. 
Finally planar segments attach boundary pixels in a similar way to such a stage for linear clusters, and this concludes the segmentation process.

Unfortunately, the implementation of the Klinker's algorithm was not preserved even by her co-authors \cite{maxwell1997physics}.
    \subsection{Nikolaev's algorithm}
 
Nikolaev's algorithm  \cite{nikolaev} is based on a model proposed by Nikolayev that is more general than the DRM model of the spectral distribution $F(\lambda)$ of the illumination of the sensor, recording the image with complex lighting conditions (more than one light source) \cite{brill1990image} and containing objects with different reflective surface properties, including metals. 
In this model the $F(\lambda)$ of each point of the object is an element of the linear submanifold of dimension $r$ (which is called the rank of the object) in the spectral function space.

The sensor projects infinite-dimensional spectral illumination distributions into a three-dimensional RGB CSp, and this preserves the degeneracy if $r < 3$: the points of the object form clusters in the CSp, lying in the same plane ($r=2$), on a straight line ($r=1$) or being a point ($r=0$).
In some cases the cluster in the CSp may have a rank not equal to, but lower than the rank of the object. 
Note that the ranking of sets in the CSp corresponds exactly to the classification (pointlike, linear, planar or volumetric) used in the Klinker's algorithm, but Nikolayev applies it to clusters corresponding to the whole objects, while Klinker classifies only areas of the preliminary segmentation.

The authors of \cite{nikolaev} described a set of scenes containing objects with different types of surfaces and located in various lighting and observation conditions, creating clusters of different ranks. In particular, flat objects illuminated by a distant light source, observed from afar, have a rank 0 according to the linear theory. 
Strongly matted and/or metallic objects (the reflectance model for which is proposed by Healey and Tominaga \cite{healey1989using,tominaga1994dichromatic}), when illuminated by one close source, have a rank 1. 
Convex glossy chromatic dielectrics, when illuminated by a close source, have a rank 2. 
In addition, rank~2 have convex glossy chromatic dielectrics illuminated by two sources: close chromatic and diffuse ones. 
An extended example list of scenes of various ranks for 3D CSp is given in table 1 of \cite{nikolaev}. 
Thus, it is possible to reformulate the problem of CS in the following form: in order to segment the image, one needs to decompose the colour histograms into point-like, linear and planar distributions. 

Nikolaev's algorithm has two stages.
At the first stage, the Gaussian filtering of the image is carried out, followed by pre-segmentation using morphological watersheds \cite{vincent1991watersheds}. 
The second stage is the stage of main segmentation, where the region adjacency graph (RAG) \cite{tremeau2000regions} is constructed and the region merging technique is consistently applied on this graph with three different weight functions.
The weight functions are chosen under the assumption that all segments have a rank of 0, 1, or 2, respectively, but the algorithm as a whole provides segmentation of images containing arbitrary combinations of segments of all listed ranks.  
We note that region-based image segmentation methods are often used in development of automatic segmentation systems~\cite{budzan2018automated, haris1998hybrid}.

The algorithm takes into account the fact that region merging with a weight function of a higher rank breaks the boundaries of objects of lower rank since it is always possible to draw a plane through a point and a straight line.  
To avoid this, the segments corresponding to uniformly coloured areas of the scene objects (with some degree of certainty) are excluded from processing (i.e. marked isolated) between stages of merging with different weight functions.

Unfortunately, the original implementation of the Nikolaev's algorithm was not preserved, but later it was partially re-implemented again \cite{khanipov}.  
In the new implementation the pre-segmentation phase was eliminated, and individual pixels were used as the initial partitioning elements. 
Khanipov's implementation does not involve segmentation of objects ranked other than 0, and contained only one region merging cycle. It was further extended for the colour-texture segmentation \cite{kunina2018aerial}. 
    \subsection{Nikolaev's and Klinker's algorithm in comparison}

Nikolaev's and Klinker's algorithms are compared in table \ref{tab:comparsion}.  
As we see, Nikolaev's algorithm is based on a more complete colour model and uses a well-researched graph-based region merging technique as an infrastructure, while  Klinker's algorithm uses a number of unique heuristics reflecting the knowledge about the shape of clusters in the CSp, which improves the accuracy of segmentation. 
It comes to mind that Nikolaev's algorithm modification with techniques similar to the one used in Kinker's algorithm would be beneficial. 

\begin{table}[h]
    \begin{center}
    \caption{\label{tab:comparsion} Comparison of Klinker's and Nikolaev's algorithms.}
    \begin{tabular}{| p{0.32\linewidth} | p{0.32\linewidth} | p{0.32\linewidth} |} 
        \hline 
         & Klinker's & Nikolaev's \\        
        \hline 
        Availability of the implementation & Lost & Lost but reconstructed \\
        \hline
        Optical image formation model & Does not take metals into account, allows only a single light source &
        Takes metals into account, allows multiple light sources  \\
        \hline 
        Additional heuristics & Consideration of L-shaped clusters of rank 2 for highlights, deep shadows and the off-scale area analysis & No \\
        \hline 
        Algorithm infrastructure & A complex set of actions &  Greedy merging technique supplemented by the edges locking \\
        \hline 
        Use of region-competition to improve the accuracy of segmentation & Partly & No \\
        \hline 
    \end{tabular}
    \end{center}
\end{table}
    
    \section{Proposed Algorithm for Colour Segmentation} 

In this work we propose a CS algorithm based on Nikolaev's approach supplemented with heuristics based on Klinker's observations. The proposed modifications are the following:

\begin{enumerate}[nolistsep]
\item Bilateral filter is used for image pre-processing. 
\item Individual pixels are used as elements of initial segment set as in \cite{khanipov}.
\item RAG edges costs and algorithm termination criteria are now derived from the general approach to minimise the sum of squared deviations of the image from its linear model (section \ref{weight_func_approach}). 
\item The algorithm is working in a projectively transformed CSp, not in the linear CSp of the sensor (section \ref{colour_space_proj_transform}).
\item L- or T-shape of rank 2 clusters is taken into account (section \ref{rank2_L_clusters}).
\item Geometric heuristic to include off-scale (overexposed or over-saturated) areas into regions is used (section \ref{offscale}).
\end{enumerate}


    \subsection{Approach to edge weight function construction} 
\label{weight_func_approach}

Segment merging in Nikolaev's algorithm is governed by the weight functions estimating colour proximity of two segments. In this work we propose an approach to weight function construction that is general for all ranks and is based on 
minimisation of sum of squared deviations (SSD) of image pixel values from segments models.

Within the linear colour theory we consider 3 models of segments corresponding to ranks from 0 to 2: point, straight line and plane. We assume that all segments have the same rank $r$ on each iteration of the region merging algorithm. Here, for each segment we would choose parameters $I$ for the model with the rank $r$ (i.e. for point, straight line or plane) in order to minimise the SSD of pixels of the segment from the model. 

Suppose at some iteration of region merging for rank $r$ we get a segmentation  $S^M = \{S_m\}_{m \in \{1, .., M\}}$ of image into $M$ segments. Let each segment $m$ be given its model $I^M_m$ with rank $r$. Let $\rho_r(\vec {I^M_m}, \vec p)$ be the distance from pixel $p \in S_m$ to model $I^M_m$ in CSp. Parameters of such a model are estimated with the least-squares method.

In assumption of the linear image formation model and normal noise lets define the cost function of fragmentation $S^M$ as the sum of squared deviations of pixels from models of their corresponding segments:
\begin{equation} \label{eq:2sum}
    U(S^M) = \sum_{m=1}^M \sum_{i=1}^{n_m} \rho_r^2 (I^M_m, \vec{p}_{m, i}), 
\end{equation}
where $\vec{p} = (\mathrm{R}, \mathrm{G}, \mathrm{B})^{\mathrm{T}}$ are coordinates of pixels in CSp and $n_m$ is the size of $m$-th segment. 

We can view the CS as the minimisation of (\ref{eq:2sum}) and approximate it with the greedy merging of the RAG.
In terms of (\ref{eq:2sum}) a neighbouring pair of segments $S_l$ and $S_k$ \mbox{from $S^M$} is an optimal merging if it leads to the smallest difference $U(S^{M-1}) - U(S^M)$, where \mbox{$S^{M-1} = S^M \backslash \{S_l, S_k\} \cup \{S_l \cup S_k\}$}. Hence, the cost $d_r(k, l)$ of merging segments $S_k$ and $S_l$ could be associated with this difference and given as 
\begin{equation}\label{eq:diff}
d_r(k, l) = U(S^M) - U(S^{M-1}) = \sum_{i=1}^{n_k+n_l} \rho_r^2 (I^{M-1}_T, \vec{p}_{T, i}) - \sum_{i=1}^{n_k} \rho_r^2 (I^M_k, \vec{p}_{k, i}) - \sum_{i=1}^{n_l} \rho_r^2 (I^M_l, \vec{p}_{l, i}),
\end{equation}
where $T$ is a result of joining segments $k$ and $l$.

Note that for SSD estimation for pixels of $m$-th segment from model $\sum_{i=1}^{n_m} \rho_r^2 (I^M_m, \vec{p}_{m, i})$ it is enough to know segment area, its centre of mass and co-variance matrix of its pixels. These three characteristics can be calculated using additive statistics, namely using area and its 1 and 2 moments (i.e sum of pixel components $\sum_{i=1}^{n_m}\{\mathrm{R}_i, \mathrm{G}_i, \mathrm{B}_i\}$ and sum of all possible pairwise products of pixels components $\sum_{i=1}^{n_m}\{\mathrm{R}_i, \mathrm{G}_i, \mathrm{B}_i\} \{\mathrm{R}_i, \mathrm{G}_i, \mathrm{B}_i\}$). Thus the calculation of RAG edges costs with weight function (\ref{eq:diff}) may be implemented without iterating through all segment points and  requires $O(1)$ time.


Such approach to RAG edge weighing is well-known \cite{koepfler1994multiscale} and is used, for example, in SAR image segmentation\cite{redding1999efficient, shui2014fast}, but applied to CS presumably for the first time. Note that in \cite{crisp2002fast} as well as in original Nikolaev's algorithm merging terminates when the weight of the best edge exceed a given threshold. Such termination criterion seems natural, but does not have a reasonable mathematical proof. Firstly, the weight of the merged edge have the meaning of error increment, not the full error, and so cannot be easily connected with expected noise level of the image. Secondly, for the incomplete graph (and the RAG is a planar graph, i.e. almost always incomplete) monotonous growth of the weight cannot be guaranteed in the sequence of the best edges. So we propose another criterion.


On every step of merging we estimate the sum of squared deviations $U(S^{M-1})$. As follows from (\ref{eq:diff}), this can be done with one addition operation $U(S^{M-1}) = U(S^M) + d_r(k, l)$.
Merging terminates when the value $\sqrt{\frac{U(S^M)}{N}}$, where $N$ is the total amount of image pixels, exceeds a given threshold.
When merging with the weight function $d_0$, $\sigma_0$ -- an algorithm parameter proportionate to the noise -- is used as the threshold. When merging with weight functions $d_1$ and $d_2$, $\sigma_1 = \sqrt{\frac{2}{3}}\sigma_0$ and $\sigma_2 = \sqrt{\frac{1}{3}}\sigma_0$ are used as thresholds.  

    \subsection{Colour space projective transform} 
\label{colour_space_proj_transform}

Note that approach with the SSD minimisation introduced above is still applied when making a transition to the transformed CSp in a case when such transformation does not change linear properties of clusters models, i.e rank structure is persisted. This requires one-to-one correspondence of planes of initial space to planes of the target space and, as a result, one-to-one correspondence of lines and points of initial space to  lines and points of the target space respectively. In other words, it is possible to apply CS in projectively transformed CSp. Homography of 3-dimensional CSp proposed for the first time presumably in \cite{gong20173d}, but applied to another problem -- photo-realistic colour transfer.


Besides the rank classification in the linear colour theory there is a sub-rank classification of clusters in CSp \cite{nikolaev}. It considers cluster orientation relatively to coordinate axes of CSp, what allows to reduce the dimension of space for clusters analysis in most cases. The requirement of sub-rank structure persistence when using CSp homography impose some additional restrictions on it:

\begin{itemize}[nolistsep]
    \item origin of initial space should map into origin of the target space, 
    \item the line through points $(0,0,0)$ and $(1,1,1)$ of~the initial space should map into the line through the same points of~the target space. 
\end{itemize}


Further restrictions to CSp homography will be introduced according to the following. 

Let us consider two pixel pairs $(\vec{p}_a, \vec{p}_b)$ and $(\vec{p}_b, \vec{p_c})$, so $\norm{\vec{p'}_a - \vec{p'}_b} = \norm{\vec{p'}_b - \vec{p_c'}}$ and $l_c > l_a$, where $l=\frac{R + G + B}{3}$ is a pixel brightness. Then we will distort the CSp so $\norm{\vec{p'}_a - \vec{p'}_b} < \norm{\vec{p'}_b - \vec{p_c'}}$. In the context of merging with region adjacency graph such space transformation will result in edges weight increasing the more, the less is the average brightness of pixels segments, corresponding to edges end-points. Such approach can be used for solving conflicts when merging segments near the zero brightness. Here conflicts occur since of all matte clusters merge near the dark corner of the colour cube, what was specified by Klinker in \cite{klinker1990physical}, where she excludes dark-colour pixels from colour analysis.

Let us require homography to be symmetrical with respect to the brightness axis (i.e line through the $(0, 0,0)$ and $(1,1,1)$). Note that although the non-linearity correction of colour variation in the RGB-space may improve clusters partition in the CSp by its colour from human perception point of view, it is rational to apply it independently from this transformation and CS as such (as well as the sensor transfer function non-linearity compensation), since such correction requires an information about characteristics of sensor colour coverage.

Also note that homographies of CSp differing only in scale of transformed space are equivalent. Changes in scale do not influence mutual orientation of clusters in CSp, so its only entails the necessity of recounting the merging threshold value when using segmentation with the region merging technique to achieve equivalent segmentation result.

\begin{figure}[h]
    \begin{center}
    \begin{minipage}[h]{0.32\linewidth}
        \center{\includegraphics[width=0.98\linewidth]{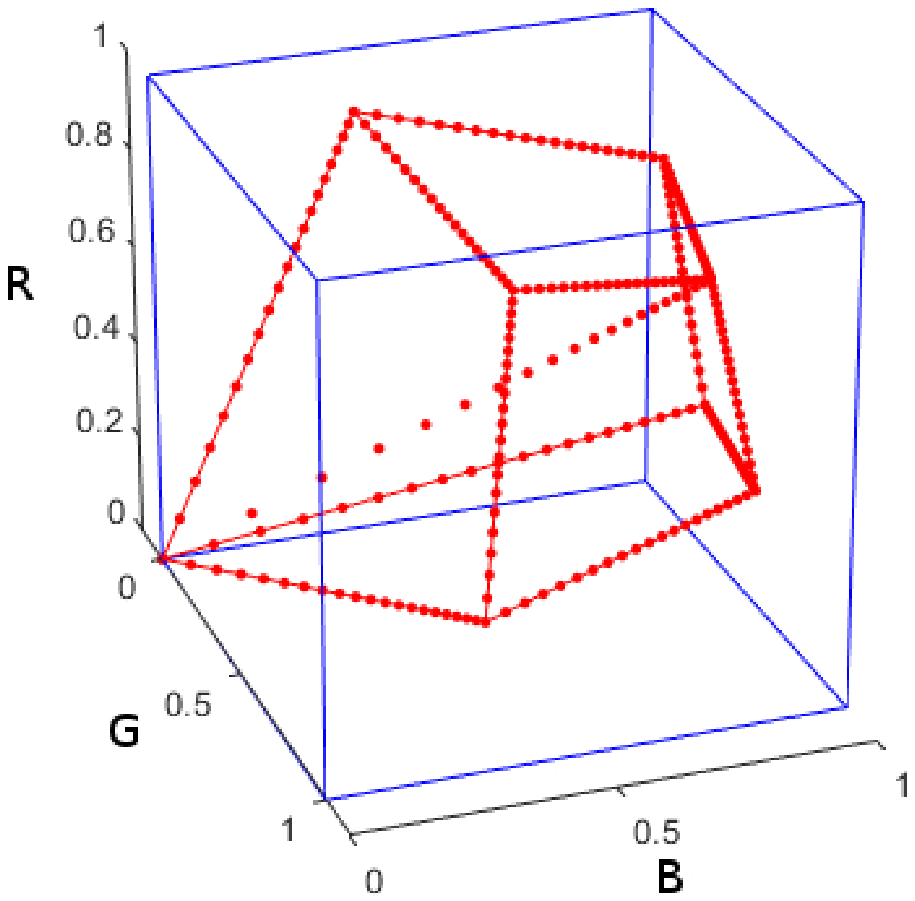} \\ a)}
    \end{minipage}
    \begin{minipage}[h]{0.32\linewidth}
        \center{\includegraphics[width=0.98\linewidth]{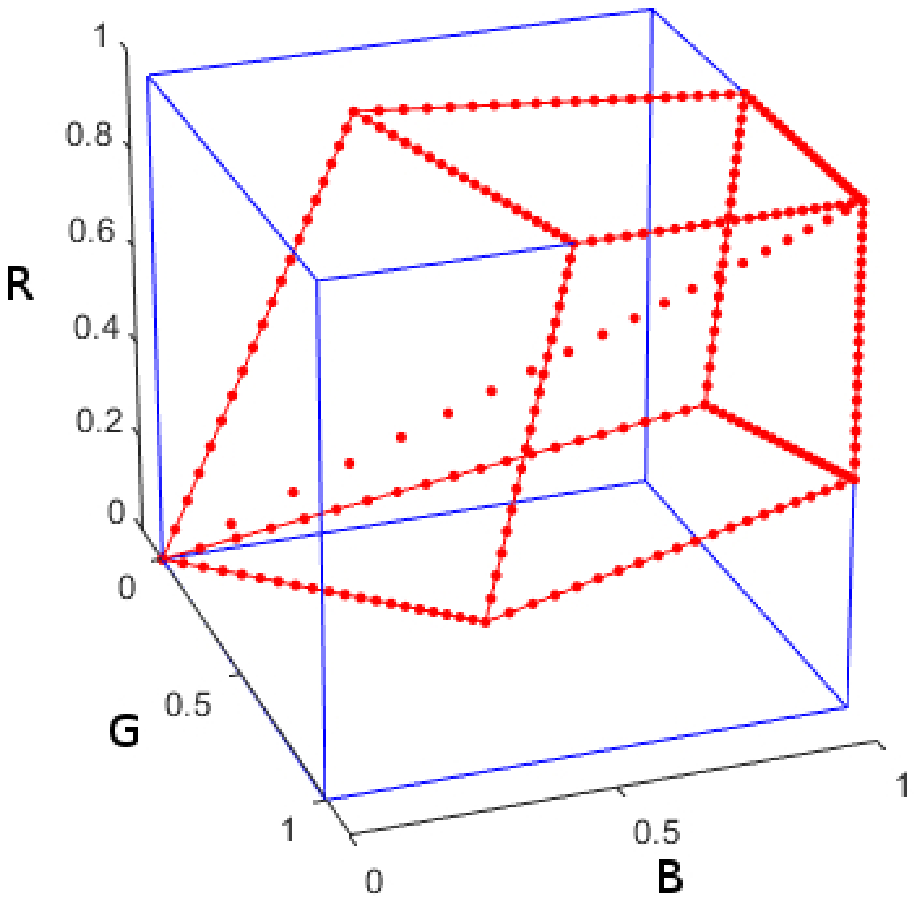} \\ b)}
    \end{minipage}     
    \begin{minipage}[h]{0.32\linewidth}
        \center{\includegraphics[width=0.98\linewidth]{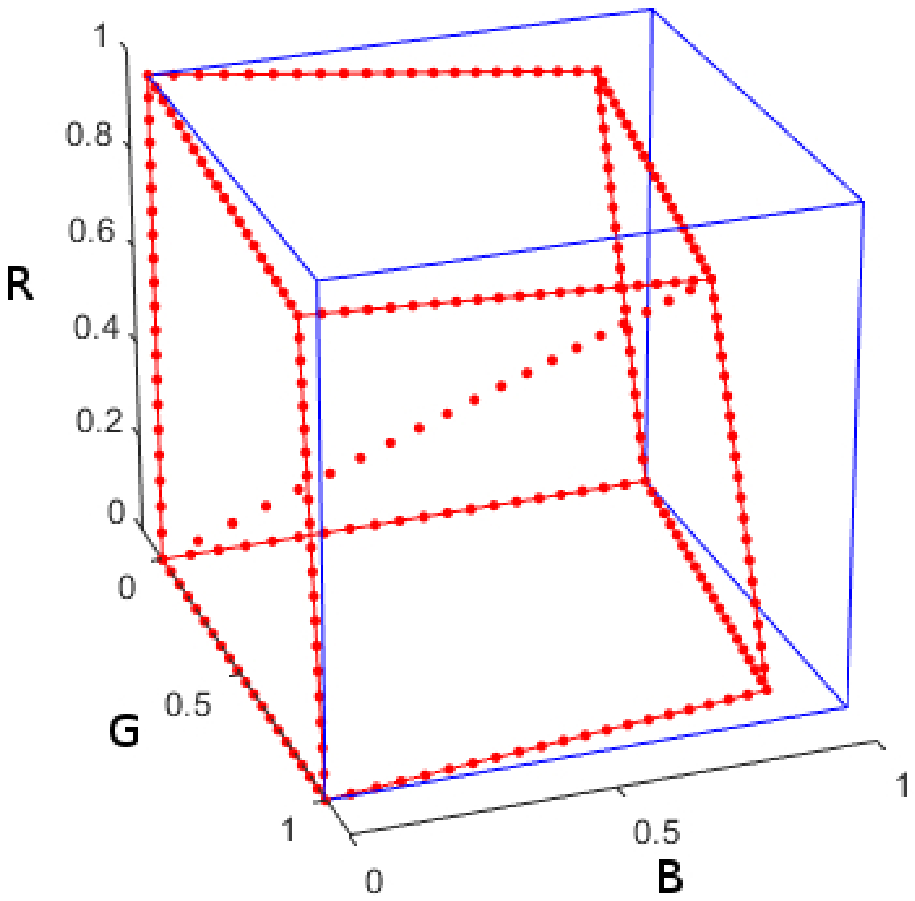} \\ c)}
    \end{minipage} \\
    \vspace{2ex}
    \caption{\label{fig:projectiveColorspaceTransform}Example of desired mapping of points of linear CSp (borders are blue) to projectively transformed CSp (borders are red) in general (a), in the particular case, when only points of the colour cube of maximum saturation are transferred (b), and in the particular case, when only compression along brightness axis of the colour cube diagonal occurs (с).} 
    \end{center}
\end{figure}

Transformation, which satisfies linear colour theory and considerations above, could be defined by the following relations of points of the initial and the target CSp:
\begin{itemize}[nolistsep]
    \item $(0,0,0) \leftrightarrow (0,0,0)$,
    \item $(1,1,1) \leftrightarrow (b,b,b)$,
    \item $(1,0,0) \leftrightarrow (1,a,a)$,
    \item $(0,1,0) \leftrightarrow (a,1,a)$,
    \item $(0,0,1) \leftrightarrow (a,a,1)$,
\end{itemize}
where $0 \leq a \leq 1$, $\frac{2a + 1}{3} \leq b \leq 1$ and $0 \leq \mathrm{R}, \mathrm{G}, \mathrm{B} \leq 1$. 
Such transformation leads to lower resolution in brightness close to point $(1, 1,1)$  of CSp, and vice versa, close to origin the resolution in brightness increases (fig.~\ref{fig:projectiveColorspaceTransform}).
When $a > 0, b = 1$ only transfer of colour cube points of maximum saturation occurs (fig.~\ref{fig:projectiveColorspaceTransform}b).
When $a = 0, b < 1$ only compression along brightness axis of colour cube diagonal occurs. (fig.~\ref{fig:projectiveColorspaceTransform}с).
When $a = 0, b = 1$ transformation is identical.


4-dimensional homography of 3-dimensional space is defined uniquely by setting five pairs of points corresponding to points transformation. 
So for each set of parameters values $a$ и $b$ there is only one projective transformation. 
Homography defined by the matrix $\mathrm{H_{4x4}}$, transforms coordinates of the pixel from the initial CSp $\vec{p}$ to $\vec{p}'$ in the following way:
\begin{equation}
    \vec{c}'_i = H_{4x4} \vec{c}_i, \nonumber
\end{equation}
where $\vec{c} = (\mathrm{R}, \mathrm{G}, \mathrm{B}, 1)^{\mathrm{T}}$ и  $\vec{c}' = (\mathrm{R'}, \mathrm{G'}, \mathrm{B'}, \mathrm{W})^{\mathrm{T}}$  -- four-dimansional homogeneous coordinates of the vectors $\vec{p}$ and $\vec{p'}$ respectively,  
$\vec{p'}_i = (\mathrm{R}' / \mathrm{W}, \mathrm{G}' / \mathrm{W}, \mathrm{B}' / \mathrm{W})^{\mathrm{T}}$. The desired parametric transform family is defined by homography matrix as
\begin{equation}\label{eq:projectiveColorspaceTransform}
    H_{4x4} = \begin{pmatrix}b & ab & ab & 0 \\ ab & b& ab & 0 \\ ab & ab & b & 0 \\
     a - \frac{b}{2} + \frac{1}{2} & a - \frac{b}{2} + \frac{1}{2} & a - \frac{b}{2} + \frac{1}{2} & -a + \frac{3b}{2} - \frac{1}{2} \end{pmatrix}.
\end{equation}.

    \subsection{Considering L- or T-shape of rank 2 clusters}
\label{rank2_L_clusters}
 
L- and Т-shape of rank 2 clusters, noticed by Klinker in \cite{klinker1990physical}, lies beyond the linear theory. In segmentation algorithm we consider L- or T-structure of two rank 1 clusters union as an additional check before applying region merging technique with the weight function (\ref{eq:diff}) for rank 2. 
In order to do that we model each of merging clusters by a segment of a straight line, which is the major axis of pixels dispersion ellipsoid in CSp. 
The centre of this segment coincides with the dispersion ellipsoid centre, and its length equals doubled square root of dispersion ellipsoid semi-axes.
We assume that two clusters forms a skew T- or L-shape, if segments modeling them intersect in the way that at least one segment has at least one end, so the distance from it to another segment is less than threshold $\delta_L$, which is provided as an input of the segmentation algorithm.

    \subsection{Geometric heuristic to include off-scale areas into regions}
\label{offscale}

The pixels in the off-scale (overexposed or over-saturated) areas generally do not fall into the planar slice defined for the dichromatic plane of an object area, as a result the planar segmentation excludes these off-scale areas.
A geometric heuristic could be applied to to include distorted pixels into the region.
As it was noticed by Klinker in \cite{klinker1990physical}, such pixels with distorted colours generally are found in the middle of a highlight region.

In the proposed algorithm, we apply this heuristic after region merging with the edge weight function $d_2$.
The regions in the RAG are considered to be off-scale if its average brightness exceeds a given threshold $\mu_B$.
If an off-scale region has only one neighbour than these two are merged.
In addition to the above-mentioned Klinker's observation, we consider that off-scales could also occurs as stripes across the body, which is a typical case for cylindrical objects.
So, if two regions adjacent to an off-scale region, but not neighbouring with each other, form an L- or T-shaped cluster, all three are also merged.

    \subsection{Formulation of proposed algorithm}

The proposed algorithm consists of the following steps:

\begin{enumerate}[nolistsep]
    
    \item Applying bilateral filtering and CSp projective transform given by (\ref{eq:projectiveColorspaceTransform}).
    
    \item Initialising RAG with vertices corresponding to individual 4-connected pixels and the SSD equal to 0.
    
    \item Applying the region merging technique with the weight function $d_0$ and a threshold $\sigma_0$ which is comparable with the noise level of the image. 
    
    \item Marking isolated segments of rank 0, i.e., those segments from which the minimal value of Kullback\mbox{-}Leibler divergence $d_{min}$ to the neighbouring segment (among all adjacent edges) exceeds a certain threshold $d_{min}~>~\sigma_G$.
    
    \item Reinitialise the SSD with a sum of squared deviations of pixel values from the models of segments of rank 1.
    
    \item Applying the region merging technique with the weight function $d_1$ and a threshold $\sigma_1 = \sqrt{\frac{2}{3}} \sigma_0 $, ignoring edges leading to isolated segments. 
    
    \item Applying the region merging technique with the weight function $d_1$ and a threshold $\sigma_1$, ignoring edges connecting two isolated segments. 
    
    \item Marking isolated those pairs of vertices that do not pass the L- or T-shape check with a threshold $\delta_L$.
    
    \item Reinitialise the SSD with a sum of squared deviations of pixel values from the models of the segments of rank 2. 
    
    \item Applying the region merging technique with weight function $d_2$ and thresholds $\sigma_2 = \sqrt{\frac{1}{3}} \sigma_0$, ignoring those edge connecting two isolated segments.
    
    \item Finally, the geometric heuristic for off-scales on a RAG with a mean brightness threshold $\mu_B$.

\end{enumerate}

Note that before the merging with the weight function $d_2$ we use the L-shape check for the rank 2 clusters with a threshold $d_L \ll \sigma_G$ to mark segments as isolated instead of using the criterion $d_{min} > \sigma_G$.


    \section{Experimental results}

\subsection{Dataset}
\label{dataset}

To evaluate the performance of the proposed algorithm we require a dataset satisfying 2 requirements. First of all, images should be recorded by a linear sensor (in order to satisfy Nikolayev's linear colour theory) or at least sensor non-linearity type and parameters should be known to compensate them. Secondly, images of scenes shouldn't contain over-detailed objects and colours which distinction is challenging even for a human. Unfortunately, datasets satisfying both requirements were not found in public sources. Therefore specific dataset was collected and released~\cite{OurDataset}. The dataset consists of three distinct parts (sub-datasets). Each part is fully acquired with the use of a single camera sensor.


The first part of the dataset consists of 19 637x468 pixels images of natural scenes selected from Barnard's DXC-930 SFU dataset for colour research \cite{SFUdatasetPaper} (further -- \textit{Selected-SFU}). 
In the original dataset each scene is taken under 11~different close light sources. 
In this work scenes under Philips Ultralume were chosen. 
Among chosen scenes 17 images contain flat varied in colour sheets of paper (so-called "mondrians"), 5 images contain volumetric objects without or with insignificant highlights, on 2 more images volumetric objects with highlights and  inter-reflection effects are depicted. For all images linear contrast adjustment with 95\% quantile was applied, 5\% of pixels were allowed to be saturated.

The second and the third parts of the dataset consist of natural scene images taken by our colleagues at IITP RAS that were used in \cite{nikolaev} but not published.
There are 5 about 450x500 images of scenes with close light source (further -- \textit{IITP-close}) and 7 1280x1024 images of scenes with diffuse light source (further -- \textit{IITP-diffuse}). The camera model for \textit{IITP-close} is  unfortunatelly unknown. Images from \textit{IITP-diffuse} sub-dataset was taken with Olympus D600L camera. All images contain volumetric objects, some of which with highlights, but with no inter-reflection effects.   

\begin{figure}[h!]
    \begin{center}
    \begin{minipage}[h]{0.3\linewidth}
        \center{\includegraphics[width=0.95\linewidth]{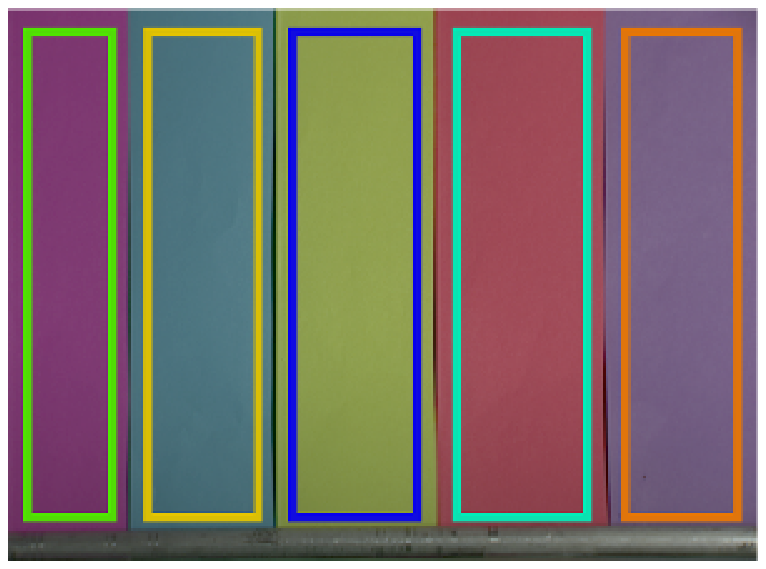}}
    \end{minipage}
    \begin{minipage}[h]{0.3\linewidth}
        \center{\includegraphics[width=0.8\linewidth]{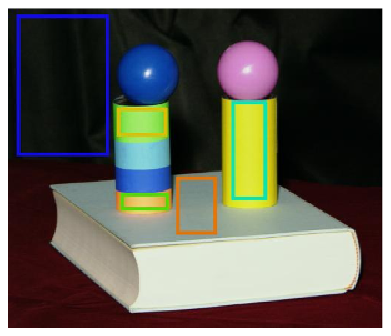}}
    \end{minipage}    
    \begin{minipage}[h]{0.3\linewidth}
        \center{\includegraphics[width=0.95\linewidth]{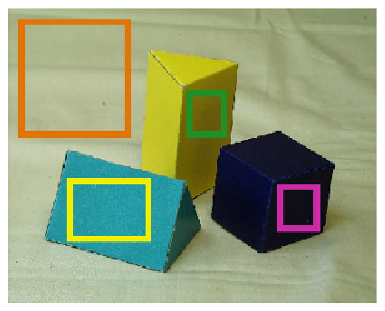}}
    \end{minipage} \\

    \begin{minipage}[h]{0.3\linewidth}
        \center{\includegraphics[width=0.95\linewidth]{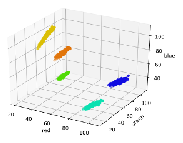} \\ Selected-SFU}
    \end{minipage}
    \begin{minipage}[h]{0.3\linewidth}
        \center{\includegraphics[width=0.95\linewidth]{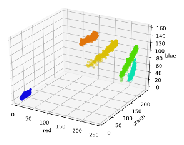} \\ IITP-close}
    \end{minipage}
    \begin{minipage}[h]{0.3\linewidth}
        \center{\includegraphics[width=0.95\linewidth]{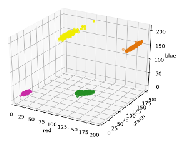} \\ IITP-diffuse} \\
    \end{minipage}    
    \vspace{2ex} 
    \caption{\label{fig:linearity} An example of rank 1 regions chosen on the test images for the examination of intensity transfer function linearity. The borders of chosen regions (top) and the corresponding clusters of pixels in the CSp (bottom) are shown with a similar colour. R, G, B components of the CSp vary from 0 to 255.}
    \end{center}
\end{figure}

To examine intensity transfer function linearity of the recording sensor, regions of rank 1 were chosen in each sub-dataset and the shape of the corresponding clusters of pixels in CSp was analysed. For all three sub-datasets parts the shape of the clusters was shown to be well approximated by a line segment (fig.~\ref{fig:linearity}).


All images are provided with pixel-by-pixel CS annotations. 
For annotation purposes, images were first automatically splitted into small subregions with guaranteed colour constancy, and then are manually merged if their colours were undistinguished by a human eye.
Segments containing less than 20 pixels were not annotated. 
In addition, annotations for regions corresponding to deep shadows (i.e. with average brightness close to zero) (fig.~\ref{fig:results}b,c) were provided. Such regions appears at \textit{Selected-SFU} images containing volumetric objects. Information about colour in such regions is lost and can not be recovered. Annotation for each shadow segment is provided as a separate binary mask.

    \subsection{Quality evaluation of proposed algorithm}

On dataset described above, we evaluated the quality of proposed algorithm, which was configured as follows. 
Bilateral filtering was applied with smoothing parameters $f_r = 50$ and $g_s = 50$, where $f_r$ is the range kernel for smoothing differences in intensities and $g_s$ is the spatial kernel for smoothing differences in coordinates. 
CSp homography was applied with parameters $a = 0$, $b = 0.4$.  Off-scale brightness threshold $\mu_B$ was tuned by an expert for each sub-dataset and provided in \ref{tab:results}. The value of $\mu_B$ is calculated for non-projectively transformed CSp in a range from 0 to 255. 

The values of thresholds on merging $\sigma_0`$ and edges locking $\sigma_G$, as well as the $\delta_L$ threshold, which is used to check L-shape of the rank 2 clusters, was chosen in a such way to achieve the best segmentation quality according to the metric. Such procedure was applied separately for each sub-dataset.

To match output segments with the ground truth ones the intersection-over-union (IoU) score, also known as Jaccard index, was calculated for each possible pair of $S^*$ ground truth and $\tilde{S}$ output segments:
\begin{equation}
    \mathrm{IoU}(S^*, \tilde{S}) = \frac{|S^* \cap \tilde{S}|}{|S^* \cup \tilde{S}|}. \nonumber
\end{equation}
which gives the ratio in [0, 1]. The evaluation quality was calculated as \textit{dataset\mbox{-}mIoU} at IoU = .50 (provides one-to-one segments matching) which is the official metric of the segmentation task in Pascal VOC \cite{everingham2010pascal} and numerous popular completions and tasks:
\begin{equation}
    \mathrm{dataset\mbox{-}mIoU} = \sum_{k=0}^{K} \mathrm{min}(\mathrm{IoU}(S^*_k, \tilde{S}_k), 0.5), \nonumber
\end{equation}
where $K$ is a number of ground truth segments. The quality range is also in [0, 1].

In a case of binary masks with shadow segments annotation are represented in dataset for a given image, output segments was firstly compared to shadow segments. 
As shadow segments may overlap each other, so, according to criterion IoU = .50 output segment may match several shadow segments. Therefore to avoid ambiguity, comparison was applied only with such shadow segment, which had maximum IoU with the output one.
Then segments unmatched with shadow segments output were compared to other ground truth segments excluding sets of pixels already matched the shadow segments.

\begin{table}[h!]
	\begin{center}
    \label{tab:results}
	\caption{Segmentation quality of proposed algorithm and optimal $\sigma_0$, $\sigma_G$, $\delta_L$ parameters configuration, the $\mu_B$ was tuned by an expert.}
	\begin{tabular}{ | c | c | c | c | c | c | } 
	    \hline 
	    Sub-dataset & $\mu_B$ & $\sigma_0$ & $\sigma_G$ & $\delta_L$ & $\mathrm{dataset\mbox{-}mIoU}$ \\
		\hline
 		Selected-SFU & 230 & 10.0 & 1.0 & 22.5 & 0.65 \\
		\hline
		IITP-close & 160 & 8.5 & 1.0 & 25.0 & 0.85 \\
		\hline
		IITP-diffuse & 250 & 6.0 & 1.0 & 30.0 & 0.71 \\
		\hline
	\end{tabular}
    \end{center}
\end{table}

\begin{figure}[!]
    \begin{center}
    \begin{minipage}[h]{0.05\linewidth}
        \hspace{5ex}
    \end{minipage}      
    \begin{minipage}[h]{0.29\linewidth}
        \center{Original image}
    \end{minipage}
    \begin{minipage}[h]{0.29\linewidth}
        \center{Ground truth}
    \end{minipage}
    \begin{minipage}[h]{0.29\linewidth}
        \center{Segmentation result}
    \end{minipage}
    \begin{minipage}[h]{0.05\linewidth}
        mIoU
    \end{minipage}   
    \vspace{0.2ex}

    \begin{minipage}[h]{0.05\linewidth}
        \center{a)}
    \end{minipage}
    \begin{minipage}[h]{0.29\linewidth}
        \center{\includegraphics[width=0.98\linewidth]{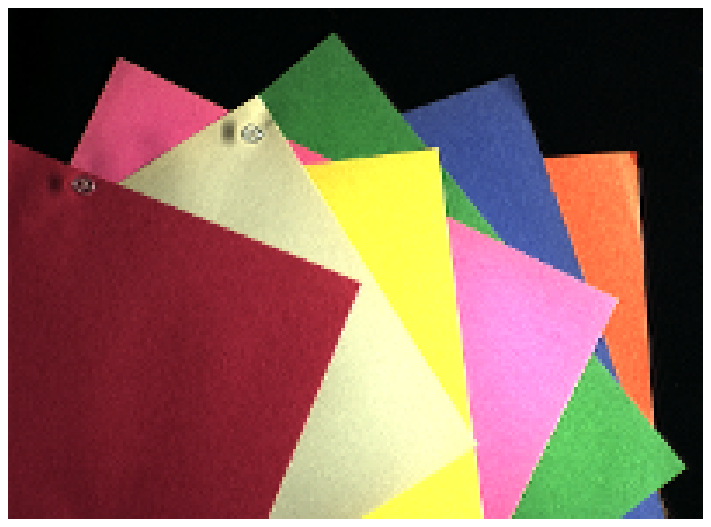}}
    \end{minipage}
    \begin{minipage}[h]{0.29\linewidth}
        \center{\includegraphics[width=0.98\linewidth]{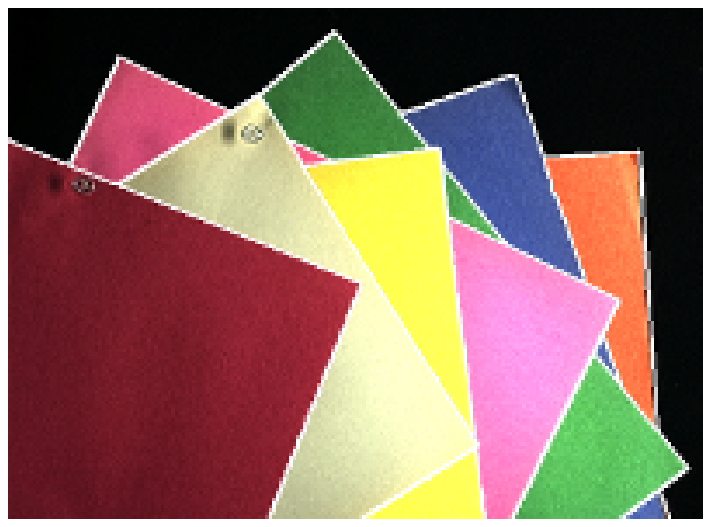}}
    \end{minipage}
    \begin{minipage}[h]{0.29\linewidth}
        \center{\includegraphics[width=0.98\linewidth]{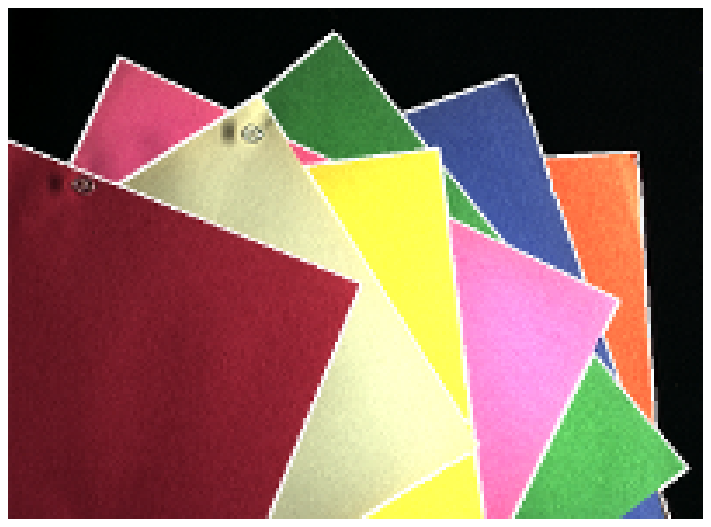}}
    \end{minipage} 
    \begin{minipage}[h]{0.05\linewidth}
        \center{0.78}
    \end{minipage}\\    
    \vspace{0.2ex}

    \begin{minipage}[h]{0.05\linewidth}
        \center{b)}
    \end{minipage}  
    \begin{minipage}[h]{0.29\linewidth}
        \center{\includegraphics[width=0.98\linewidth]{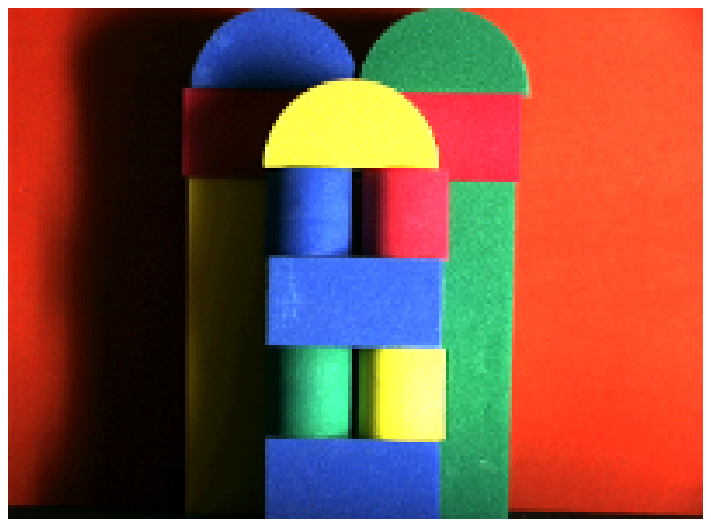}}
    \end{minipage}
    \begin{minipage}[h]{0.29\linewidth}
        \center{\includegraphics[width=0.98\linewidth]{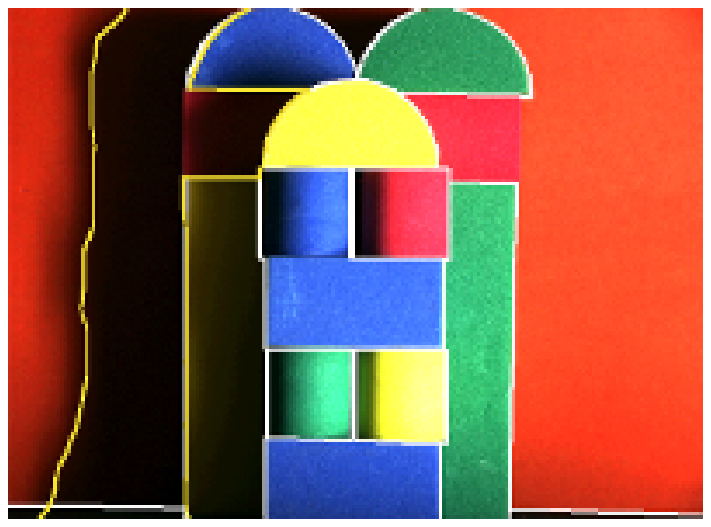}}
    \end{minipage}
    \begin{minipage}[h]{0.29\linewidth}
        \center{\includegraphics[width=0.98\linewidth]{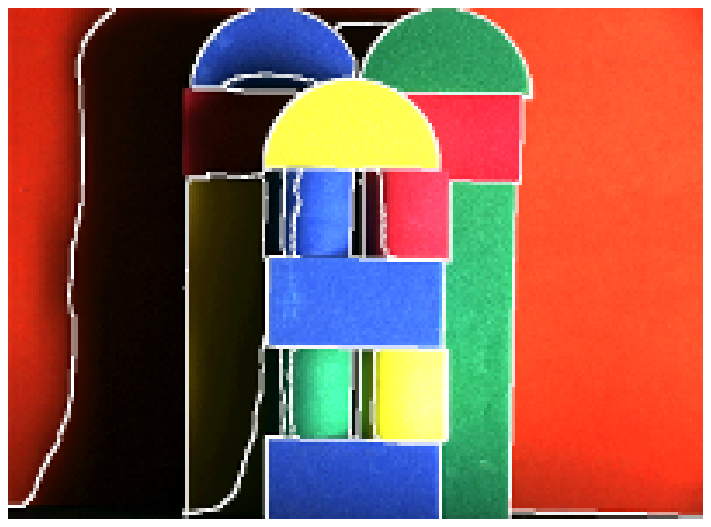}}
    \end{minipage} 
    \begin{minipage}[h]{0.05\linewidth}
        \center{0.72}
    \end{minipage}\\    
    \vspace{0.2ex}

    \begin{minipage}[h]{0.05\linewidth}
        \center{c)}
    \end{minipage}  
    \begin{minipage}[h]{0.29\linewidth}
        \center{\includegraphics[width=0.98\linewidth]{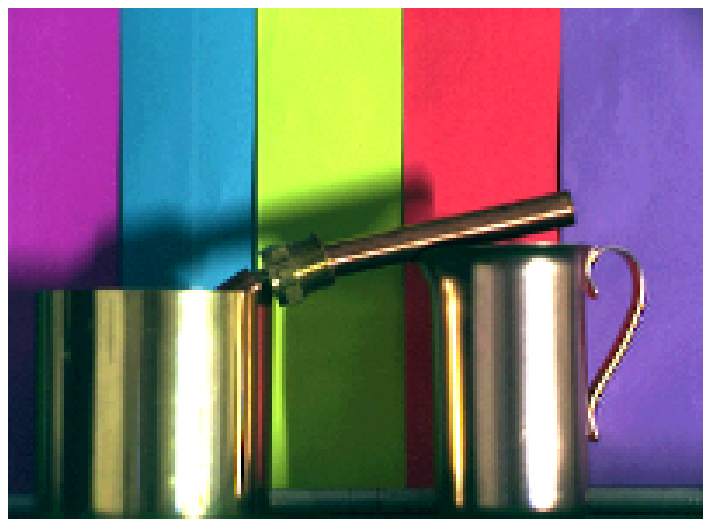}}
    \end{minipage}
    \begin{minipage}[h]{0.29\linewidth}
        \center{\includegraphics[width=0.98\linewidth]{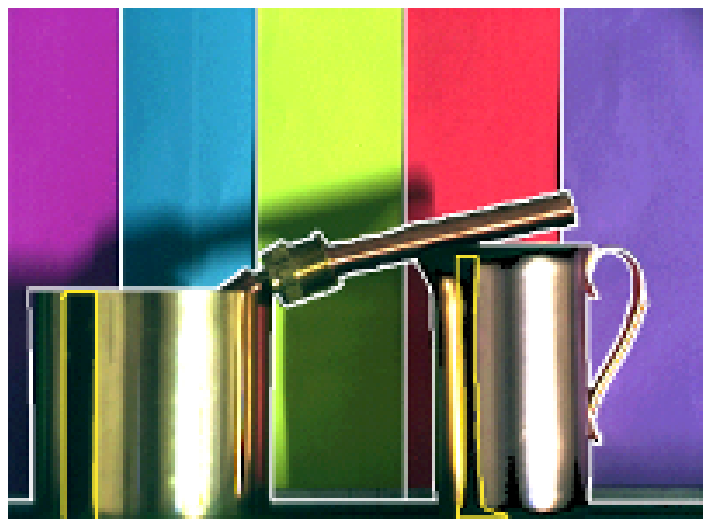}}
    \end{minipage} 
    \begin{minipage}[h]{0.29\linewidth}
        \center{\includegraphics[width=0.98\linewidth]{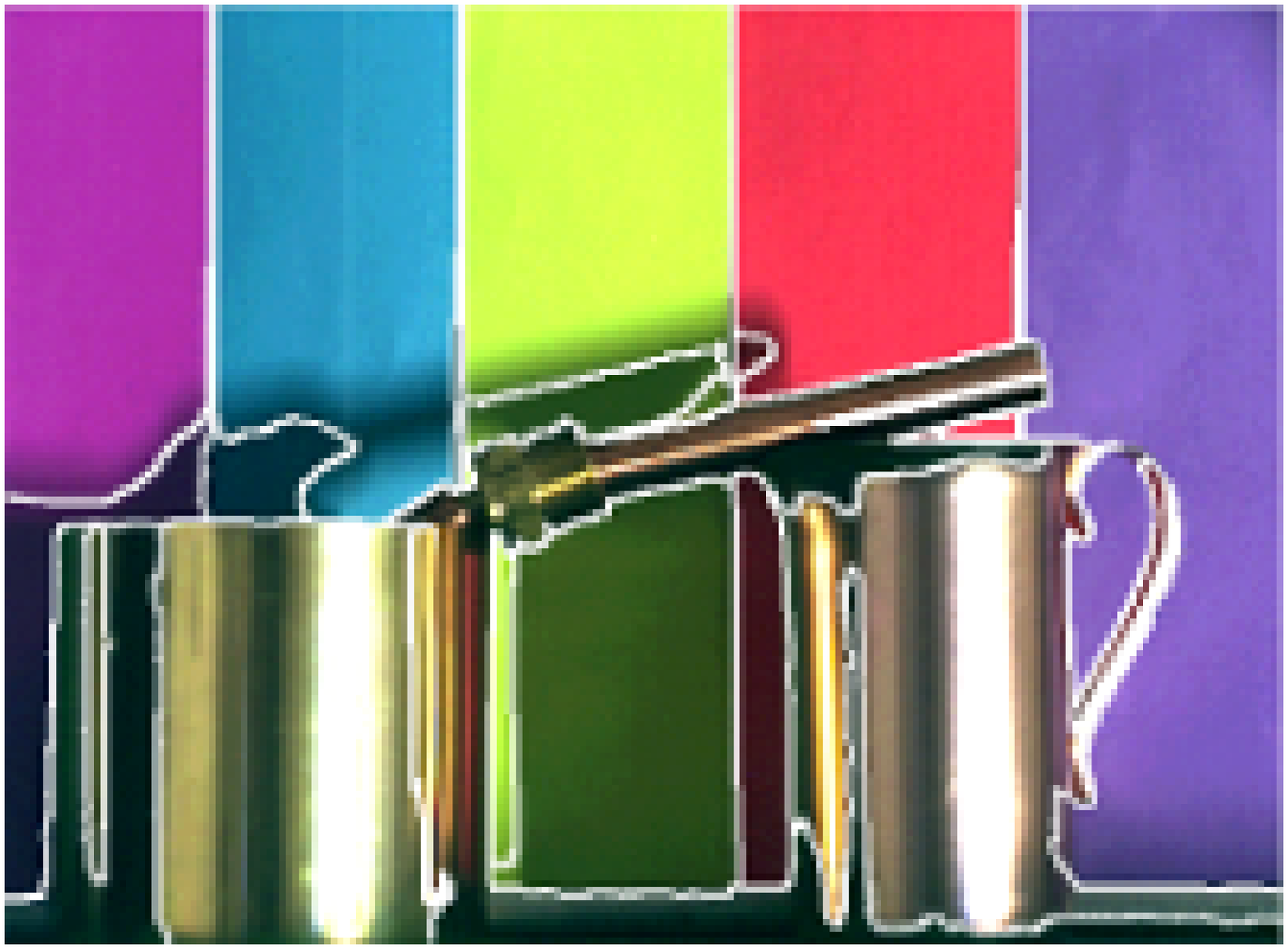}}
    \end{minipage} 
    \begin{minipage}[h]{0.05\linewidth}
        \center{0.57}
    \end{minipage}\\     
    \vspace{0.2ex}

    \begin{minipage}[h]{0.05\linewidth}
        \center{d)}
    \end{minipage}    
    \begin{minipage}[h]{0.29\linewidth}
        \center{\includegraphics[width=0.98\linewidth]{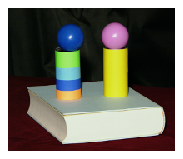}}
    \end{minipage}
    \begin{minipage}[h]{0.29\linewidth}
        \center{\includegraphics[width=0.98\linewidth]{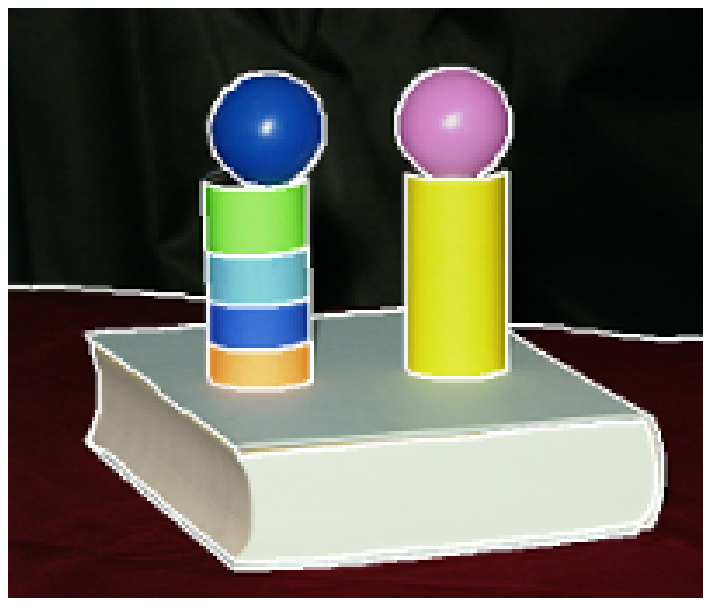}}
    \end{minipage} 
    \begin{minipage}[h]{0.29\linewidth}
        \center{\includegraphics[width=0.98\linewidth]{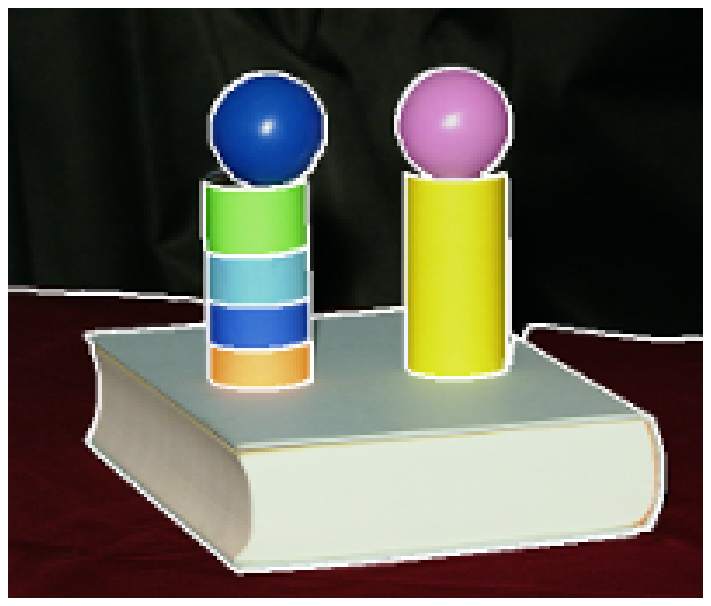}}
    \end{minipage} 
    \begin{minipage}[h]{0.05\linewidth}
        \center{0.85}
    \end{minipage}\\      
    \vspace{0.2ex}

    \begin{minipage}[h]{0.05\linewidth}
        \center{e)}
    \end{minipage}
    \begin{minipage}[h]{0.29\linewidth}
        \center{\includegraphics[width=0.98\linewidth]{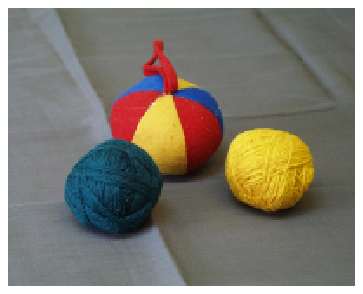}}
    \end{minipage}
    \begin{minipage}[h]{0.29\linewidth}
        \center{\includegraphics[width=0.98\linewidth]{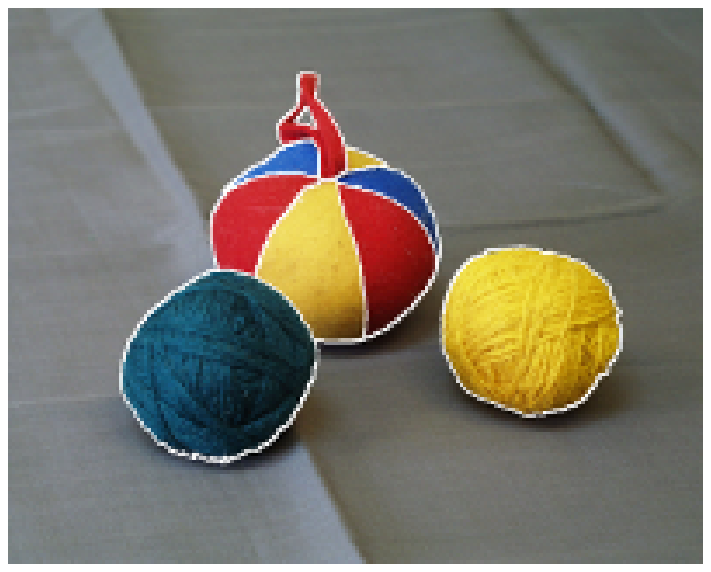}}
    \end{minipage}    
    \begin{minipage}[h]{0.29\linewidth}
        \center{\includegraphics[width=0.98\linewidth]{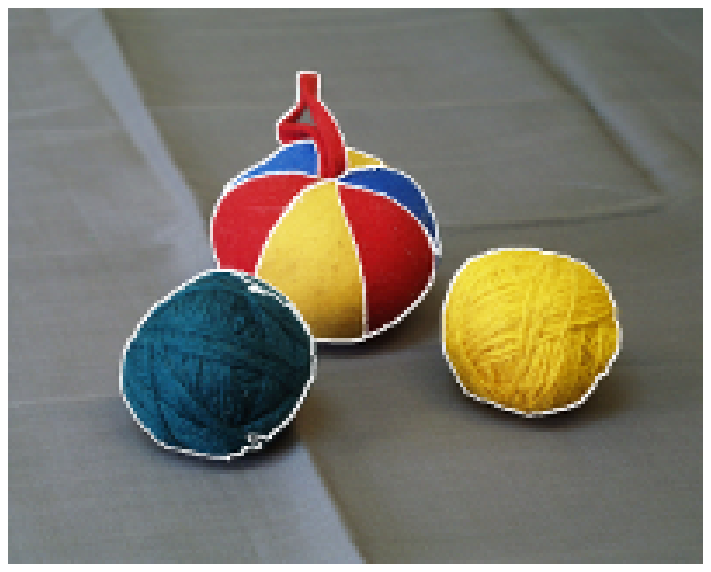}}
    \end{minipage} 
    \begin{minipage}[h]{0.05\linewidth}
        \center{0.72}
    \end{minipage}\\
    \vspace{2ex}
    \caption{\label{fig:results} Results of colour segmentation with proposed algorithm for optimal parameters configuration provided in table ~\ref{tab:results} on images from \textit{Selected-SFU} (a,b,c), \textit{IITP-close} (d) and \textit{IITP-diffuse} (e) sub-datasets. Here mIoU value is provided in correspondence to each segmented image. With yellow shadow segments are marked in the ground truth.}
    \end{center}
\end{figure}

The table \ref{tab:results} shows the experimental results for the proposed algorithm for each of the subsets of the dataset, showing the optimal values of the adjustable parameters, and the figure \ref{fig:results} illustrates the segmentation results. We see that the quality is high on the \textit{IITP-close} sub-dataset which consists of images captured in conditions similar to the model ones, whereas the quality on the \textit{Selected-SFU} subset is worst, reflecting the much more complex nature of images.

\begin{figure}[!h]
    \begin{center}
    \begin{minipage}[h]{0.24\linewidth}
        \center{\includegraphics[width=0.98\linewidth]{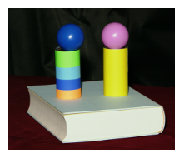}}
    \end{minipage}
    \begin{minipage}[h]{0.24\linewidth}
        \center{\includegraphics[width=0.98\linewidth]{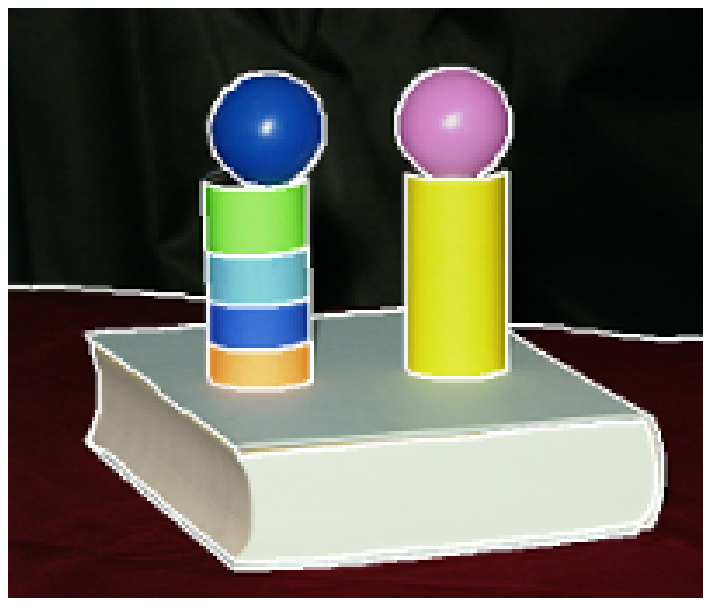}}
    \end{minipage}
    \begin{minipage}[h]{0.24\linewidth}
        \center{\includegraphics[width=0.98\linewidth]{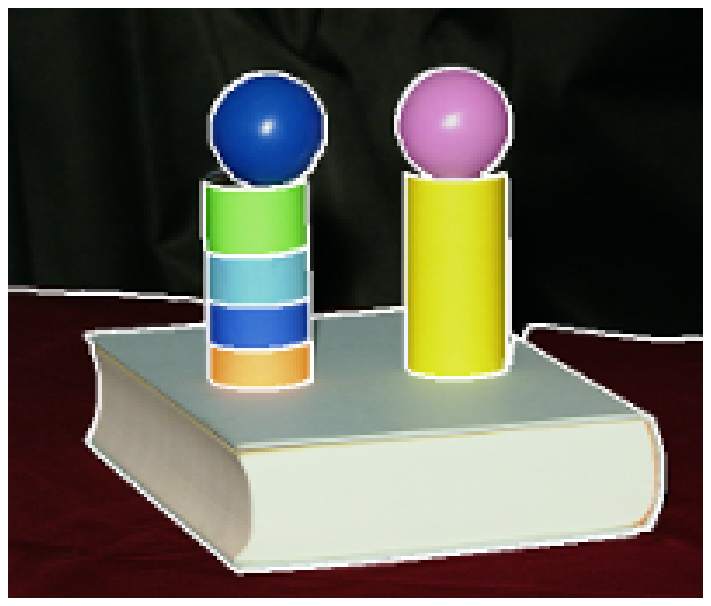}}
    \end{minipage}
    \begin{minipage}[h]{0.24\linewidth}
        \center{\includegraphics[width=0.98\linewidth]{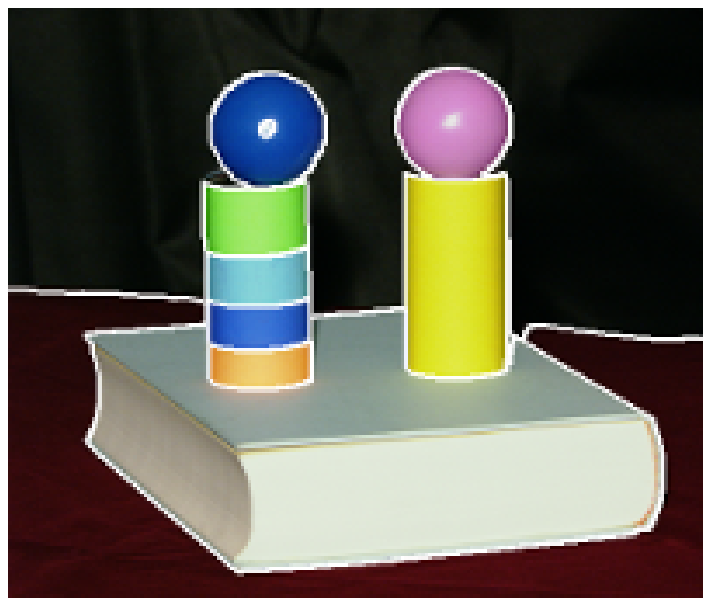}}
    \end{minipage}\\
    \vspace{0.2ex}    
    \begin{minipage}[h]{0.24\linewidth}
        \center{a)}
    \end{minipage}    
    \begin{minipage}[h]{0.24\linewidth}
        \center{b)}
    \end{minipage}
    \begin{minipage}[h]{0.24\linewidth}
        \center{c) mIoU = 0.85}
    \end{minipage}
    \begin{minipage}[h]{0.24\linewidth}
        \center{d) mIoU = 0.67}
    \end{minipage}    
    \vspace{0.2ex}    
    
    \begin{minipage}[h]{0.24\linewidth}
        \center{\includegraphics[width=0.98\linewidth]{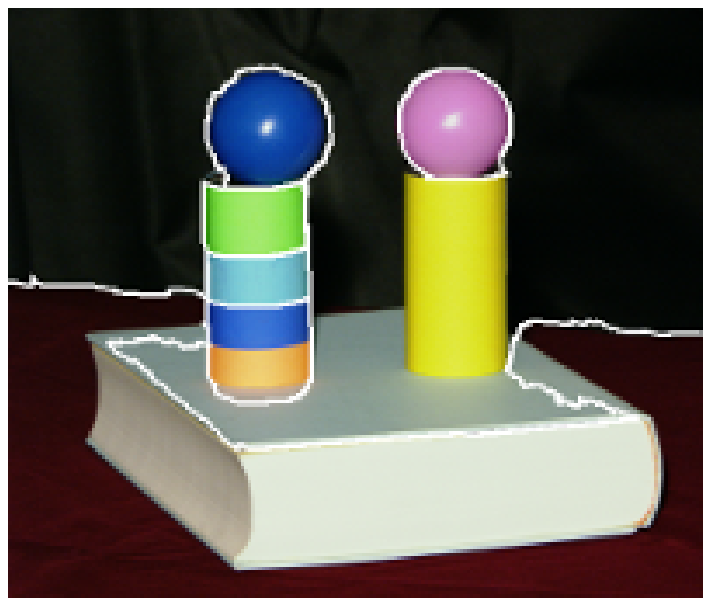}}
    \end{minipage}    
    \begin{minipage}[h]{0.24\linewidth}
        \center{\includegraphics[width=0.98\linewidth]{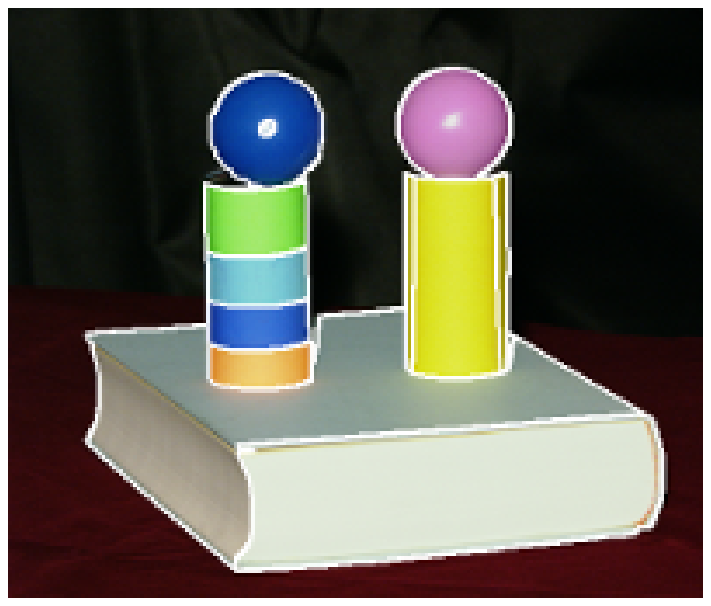}}
    \end{minipage}
    \begin{minipage}[h]{0.24\linewidth}
        \center{\includegraphics[width=0.98\linewidth]{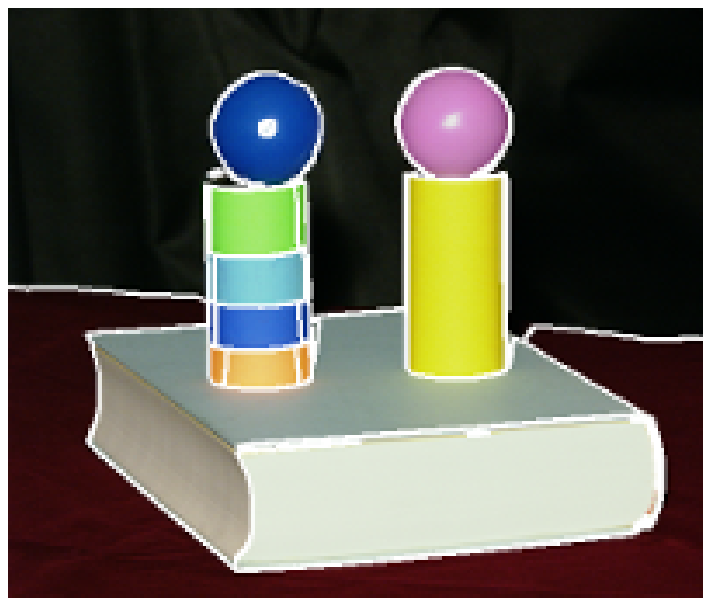}}
    \end{minipage}
    \begin{minipage}[h]{0.24\linewidth}
        \center{\includegraphics[width=0.98\linewidth]{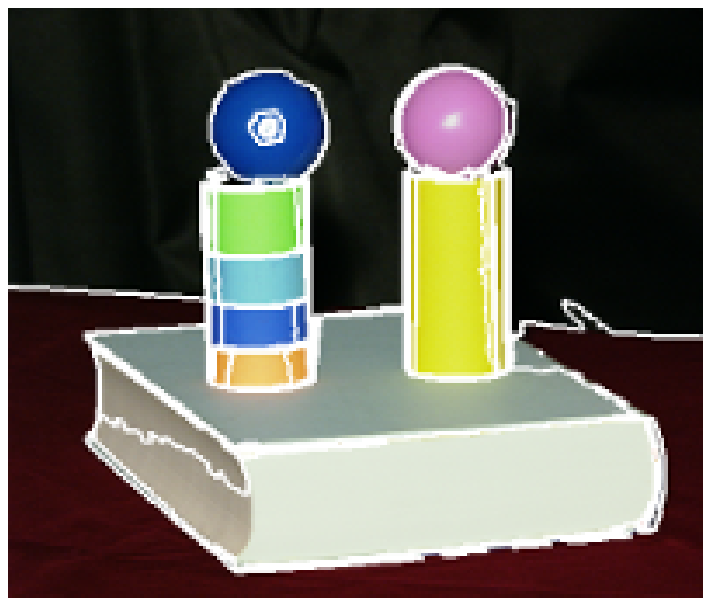}}
    \end{minipage} \\ 
    \vspace{0.2ex}    
    \begin{minipage}[h]{0.24\linewidth}
        \center{e) mIoU = 0.25}
    \end{minipage}    
    \begin{minipage}[h]{0.24\linewidth}
        \center{f) mIoU = 0.59}
    \end{minipage}
    \begin{minipage}[h]{0.24\linewidth}
        \center{g) mIoU = 0.66}
    \end{minipage}   
    \begin{minipage}[h]{0.24\linewidth}
        \center{h) mIoU = 0.61}
    \end{minipage}       
    \vspace{2ex} 
    \caption{\label{fig:comparison} How the proposed algorithm works with modifications disabled: original image (a),  ground truth (b), segmentation results with all modifications enabled in proposed algorithm (c), without geometric heuristic for off-scales (d), without cluster L- or T-shape check (e), without CSp homography (f), with old weight functions (g), with Gaussian filter ($\sigma=3$) instead of the bilateral one.}
    \end{center}
\end{figure}

Since we do not have the possibility to compare the results of the proposed algorithm with Klinker's or Nikolaev's results, we have had to resort to testing of the proposed modifications to prove their positive impact on the segmentation result. 
We were turning them off one by one, readjusting the parameter values to be optimal, and then applying testing this restricted algorithm on the dataset. As we see from the figure ~\ref{fig:comparison}d, without the geometric heuristic for off-scales one of these areas remains unmerged with the highlight one; disabling the cluster L-shape check leads to erroneous merging of areas corresponding to different colours (fig.~\ref{fig:comparison}e); in other cases, we see false separations of colours on areas differing only in brightness even in the absence of sharp shadows (fig.~\ref{fig:comparison}e--h).

    \subsection{Properties of the proposed algorithm}

Let us consider the properties of the algorithm as seen from the figure~\ref{fig:results}. We see that the uniform CS is performed correctly in the following cases:

\begin{itemize}[nolistsep]
    \item flat and volumetric dielectric surfaces with diffuse shadows (except neighbouring very small uniformly coloured areas), without jagged contour and false positive segments near the object boundaries;
    \item deep shadows like on figure~\ref{fig:results}e;
    \item overexposed areas on the uniformly coloured dielectrics (fig.~\ref{fig:results}d) and metals (fig.~\ref{fig:results}c).
\end{itemize}




Some kinds of scenes are still difficult:

\begin{itemize}[nolistsep]
    \item image areas different only in brightness are not always segmented correctly: on fig.~\ref{fig:results}d the light blue book cover is merged with its white pages, and on fig.~\ref{fig:results}b, conversely, a part of the red wall is not merged with the red square block;
    
    \item in some cases the sharp shadow boundaries are incorrectly considered to be separate objects (fig.~\ref{fig:results}c);
    
    \item some small segments are incorrectly merged with larger neighbouring ones (fig.~\ref{fig:results}a,e), this is caused by the choice of the cost function~\ref{eq:2sum} and could be solved by its refinement;
    
    \item Some images contain the thin elongated gaps between segment boundaries (e.g. in fig.~\ref{fig:results}e between the left and central objects), this can be fixed by taking the segment perimeter into account in the cost function as it was proposed in \cite{koepfler1994multiscale}.
\end{itemize}

We specially note that our algorithm does not correctly process the inter-reflections (like in fig.~\ref{fig:results}c on the metallic pan to the left), but inter-reflections can not be described in terms of linear colour theory with rank lower than 3, and thus this behaviour is expected. We also cannot expect the correct segmentation of very deep shadows with near-zero brightness (fig.~\ref{fig:results}b) and is accounted for in the reference dataset annotation.


    \section{Conclusion}


The development of the linear CS algorithm based on Nikolaev's approach with modifications inspired by Klinker's heuristics is presented. A novel generalised approach to weight function construction is used, which is based on minimisation of the sum of squared deviations of the image from its linear model. Another proposed feature is the CSp projective transform as a pre-processing step that allows better separation of segments in low illumination areas thus better accounting for shadows, while preserving the linear properties of clusters models in the CSp.

The proposed algorithm is tested on a novel dataset which is partially based on the (rather complex) Barnard’s DXC-930 SFU dataset~\cite{SFUdatasetPaper} supplemented by images representing simpler conditions allowing the better study of the adherence of the algorithm to the considered cases of the linear colour model. The per-pixel annotation is provided for each image of the dataset. Using this dataset, we show that all the proposed modifications do in fact enhance the segmentation quality. The experimentally discovered properties of the algorithm include good processing of strong shadows and overexposed areas, while some of the drawbacks may be attributed either to limitations of the model (segment rank less than 3) or of the weight function, which may be addressed in future work. Both algorithm implementation and the dataset are available for public use at \cite{OurCode}.


    \acknowledgements

This work is partially supported by Russian Foundation for Basic Research (project 17-29-03514).

Authors thank Dr. Pavel Chochia for dataset images collected at IITP RAS. Authors also express their sincere gratitude to Dmitry Sidorchuk, Ivan Konovalenko and Alexey Glikin for their technical help.

    \bibliographystyle{spiebib}

\end{document}